\documentclass[twoside,11pt]{article}

\usepackage{blindtext}

\usepackage[preprint]{jmlr2e}
\usepackage{amsmath}
\usepackage{float}
\usepackage{subfigure}

\def\q{{\boldsymbol q}}
\def\k{{\boldsymbol k}}
\def\v{{\boldsymbol v}}
\def\p{{\boldsymbol p}}
\def\P{{\boldsymbol P}}
\def\x{{\boldsymbol x}}
\def\W{{\boldsymbol W}}

\usepackage{lastpage}
\jmlrheading{23}{2024}{1-\pageref{LastPage}}{1/21; Revised 5/22}{9/22}{21-0000}{Arpit Aggarwal}

\ShortHeadings{PoPE}{PoPE}
\firstpageno{1}

\begin{document}

\title{PoPE: Legendre Orthogonal Polynomials Based Position Encoding for Large Language Models}

\author{\name Arpit Aggarwal \email arpit.aggarwal@africa.airtel.com \\
       \addr Airtel Africa Digital Labs}

\editor{}

\maketitle

\begin{abstract}%   <- trailing '%' for backward compatibility of .sty file
There are several improvements proposed over the baseline Absolute Positional Encoding (APE) method used in original transformer. In this study, we aim to investigate the implications of inadequately representing positional encoding in higher dimensions on crucial aspects of the attention mechanism, the model's capacity to learn relative positional information, and the convergence of models, all stemming from the choice of sinusoidal basis functions. Through a combination of theoretical insights and empirical analyses, we elucidate how these challenges extend beyond APEs and may adversely affect the performance of Relative Positional Encoding (RPE) methods, such as Rotatory Positional Encoding (RoPE).

Subsequently, we introduce an innovative solution termed Orthogonal Polynomial Based Positional Encoding (PoPE) to address some of the limitations associated with existing methods. The PoPE method encodes positional information by leveraging Orthogonal Legendre polynomials. Legendre polynomials as basis functions offers several desirable properties for positional encoding, including improved correlation structure, non-periodicity, orthogonality, and distinct functional forms among polynomials of varying orders. Our experimental findings demonstrate that transformer models incorporating PoPE outperform baseline transformer models on the $Multi30k$ English-to-German translation task, thus establishing a new performance benchmark. Furthermore, PoPE-based transformers exhibit significantly accelerated convergence rates.

Additionally, we will present novel theoretical perspectives on position encoding based on the superior performance of PoPE.
\end{abstract}


\section{Introduction}
Positional encoding was perhaps one of the two major ideas along with multi-headed attention (\citealp{DBLP:journals/corr/VaswaniSPUJGKP17}) which revolutionized the field of language modeling. Unlike other language modeling paradigms based on recurrent architectures such as RNNs, LSTMs, and GRUs, position information of a token is not explicitly given in transformer models but rather incorporated implicitly by generally adding (\citealp{DBLP:journals/corr/VaswaniSPUJGKP17}) or multiplying (\citealp{DBLP:journals/corr/abs-2104-09864}) by some pre-defined function with token embedding, or in some cases learned as parameter (\citealp{JMLR:v21:20-074}) of the model, along with other variations. Several explanations have been proposed to understand the role of positional encoding in attention mechanism (\citealp{chen-etal-2021-simple}; \citealp{DBLP:journals/corr/abs-2006-03654}; \citealp{NEURIPS2023_4e85362c}; \citealp{JMLR:v21:20-074}; \citealp{DBLP:journals/corr/abs-2104-09864}).
In this paper we shall expand some theoretical aspects of position encoding, discuss shortcoming of some existing methods and propose a new absolute positional encoding scheme based on orthogonal polynomials. We shall show with empirical, theoretical, and experimental evidence that using such polynomials for encoding positional information has inherent advantages over some of the existing APE and RPE methods.
We shall approach this paper as follows:

\begin{itemize}

\item Our initial focus will be on examining the limitations of prevailing methods in accurately representing positional information, particularly in high-dimensional contexts. These shortcomings lead to the provision of a biased informative prior to the model. Through empirical and theoretical analyses, we will illustrate how this bias negatively impacts the performance of certain Absolute Positional Encoding (APE) methods and minimizes the effectiveness of some Relative Positional Encoding (RPE) techniques. We will provide compelling evidence to demonstrate that the inadequacy of most APE and RPE methods in addressing this issue is primarily attributed to the selection of basis functions.
\item Next, we introduce a novel positional encoding scheme based on Legendre polynomials. We aim to illustrate how this innovative approach effectively addresses prevalent challenges encountered in current encoding methodologies. We will elucidate the advantageous mathematical properties inherent in the Legendre polynomial family, including orthogonality, distinct functional forms, three-recursion relations, and superior correlation structures, particularly evident in higher dimensions. We assert that these properties position PoPE as a promising alternative to existing Absolute Positional Encoding (APE) and Relative Positional Encoding (RPE) methods.
\item With experimental results, we shall demonstrate that transformer models with PoPE outperforms both $base$ and $big$ transformer models on $Multi30k$ English to German translation task and consistently converges much faster during training routines.
\item In conclusion, we will put forth theoretical explanations that delve into the core mechanisms of positional encoding methods. Our aim is to elucidate why the PoPE method exhibits superior performance, followed by the slightly enhanced performance of Relative Positional Encoding (RPE) techniques like RoPE, and the advantages inherent in alternative paradigms such as learned positional encoding.

\end{itemize}

\section{Prelude to PoPE}

In this section we shall lay out the mathematical structure of position encoding and linear attention mechanism, then we shall demonstrate some shortcoming of the existing approaches through empirical evidence and mathematical analysis. Application of proposed PoPE methodology is not limited to any particular model architecture, in this paper we have chosen original transformer architecture as a case study. Later, we shall show that the principles discussed here have wider scope.

\subsection{Background and Framework}
We shall adapt mathematical notations from \citet{DBLP:journals/corr/abs-2104-09864} regarding attention mechanism of transformer models.\\
Let $\mathbb{S}_N=\{w_i\}_{i=1}^{N}$ be a sequence of $N$ input tokens.\\
 $w_i$ the $i^{th}$ token. \\
$\mathbb{E}_N = {\{\x_{i}}\}_{i=1}^{N}$ denoted the token embedding of $\mathbb{S}_N$,\\
where $\x_{i}\in\mathbb{R}^{d}$ is the d-dimensional token embedding vector of token $w_{i}$.\\
In self attention, Key, query, and values ($\q_n$, $\k_{n}$,  and $ \v_{n}$) in self attention incorporate the $m^{th}$ and $n^{th}$ positions through transformations $f_{q},f_{k}$ and $f_{v}$ (Equation~(\ref{fn:qkv})).
\begin{equation}
	\begin{aligned}
		\q_{m} &=f_q(\x_m, m)\\
		\k_n &=f_k(\x_n, n)\\
		\v_n &=f_v(\x_n, n),\\
	\end{aligned}
	\label{fn:qkv}
\end{equation}
In typical absolute positional encoding and attention scheme, positional encoding is added to the token embedding and then transformed using linear operators, such as different projection matrices for key, query, and values (Equation~(\ref{fn:adtv-posi})).
\begin{equation}
	f_{t:t\in\{q,k,v\}}(\x_i,i):=\W_{t:t\in\{q,k,v\}}(\x_i+\p_i),
	\label{fn:adtv-posi}
\end{equation}

Key and Query values are used to compute attention $a_{mn}$ using inner product of $\q_m$ and $\k_n$. The output $\mathbf{o_m}$ is the sum of value representations $\v_n$, weighted by calculated attentions (Equation~(\ref{eq: attention mechanism})).
\begin{equation}
	\begin{aligned}
		a_{m,n}&=\frac{\exp(\frac{\q_m^{\intercal}\k_n}{\sqrt{d}})}{\sum_{j=1}^{N}\exp(\frac{\q_m^{\intercal}\k_j}{\sqrt{d}})}\\
		\mathbf{o}_m&=\sum_{n=1}^{N}a_{m,n}\v_{n}
	\end{aligned}
	\label{eq: attention mechanism}
\end{equation}

The inner product of query and keys form the basis for the above attention machinery (Equation~(\ref{eq inner product attention})).

\begin{equation}
	\q_m^{\intercal}\k_n=\x_m^{\intercal}\W_q^{\intercal}\W_k\x_n+\x_m^{\intercal}\W_q^{\intercal}\W_k\p_n+\p_m^{\intercal}\W_q^{\intercal}\W_k\x_n+\p_m^{\intercal}\W_q^{\intercal}\W_k\p_n
	\label{eq inner product attention}
\end{equation}

\subsection{Empirical Analysis}
\begin{figure}[H]
    \begin{center}
    \subfigure[Cosine positional encoding]{
    \includegraphics[height=5cm,width=0.5\textwidth,scale=1]{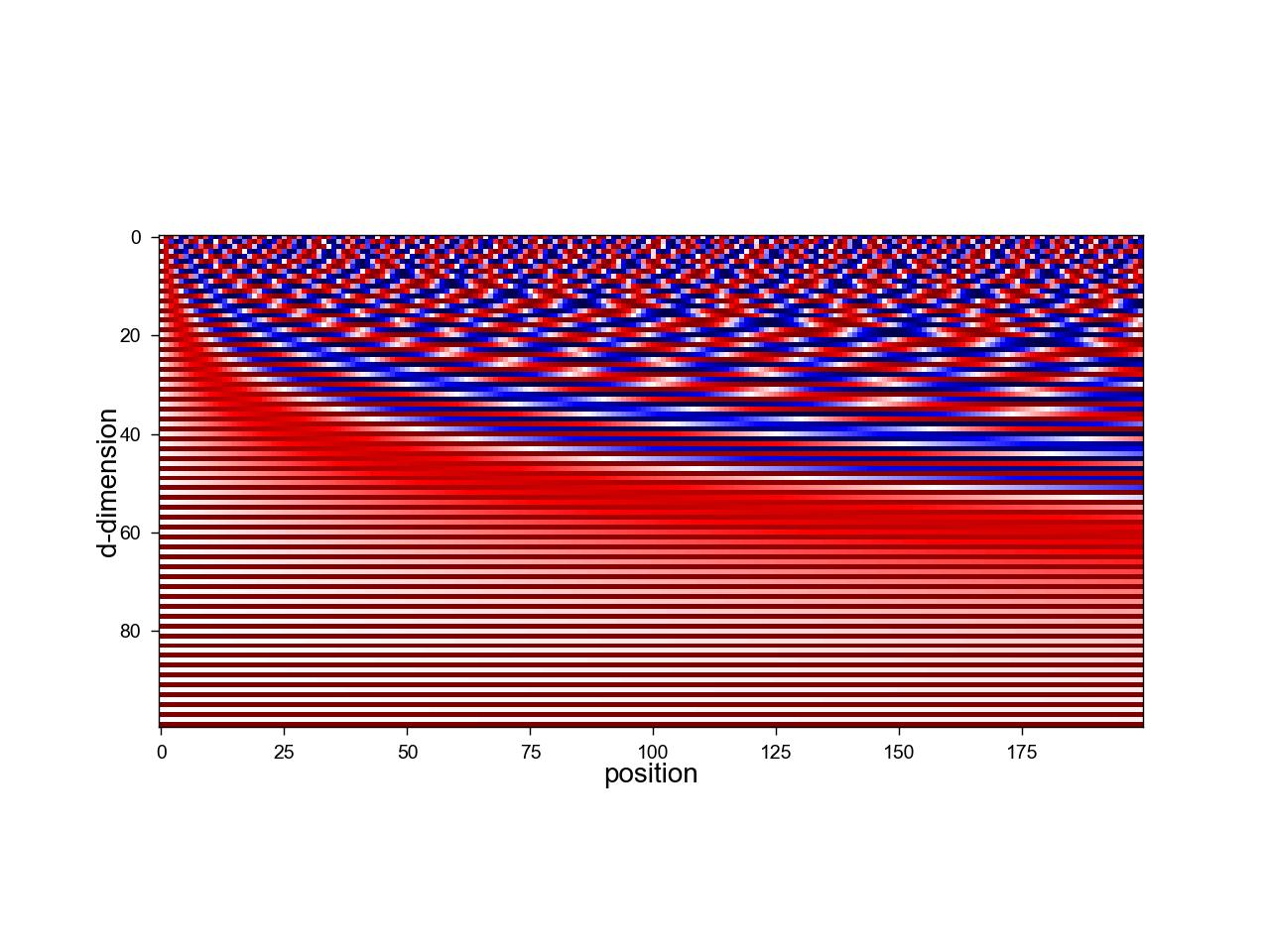}
    }
    \subfigure[Near perfect correlation at higher dimensions]{
    \includegraphics[height=5cm, scale=1,width=0.5\textwidth]{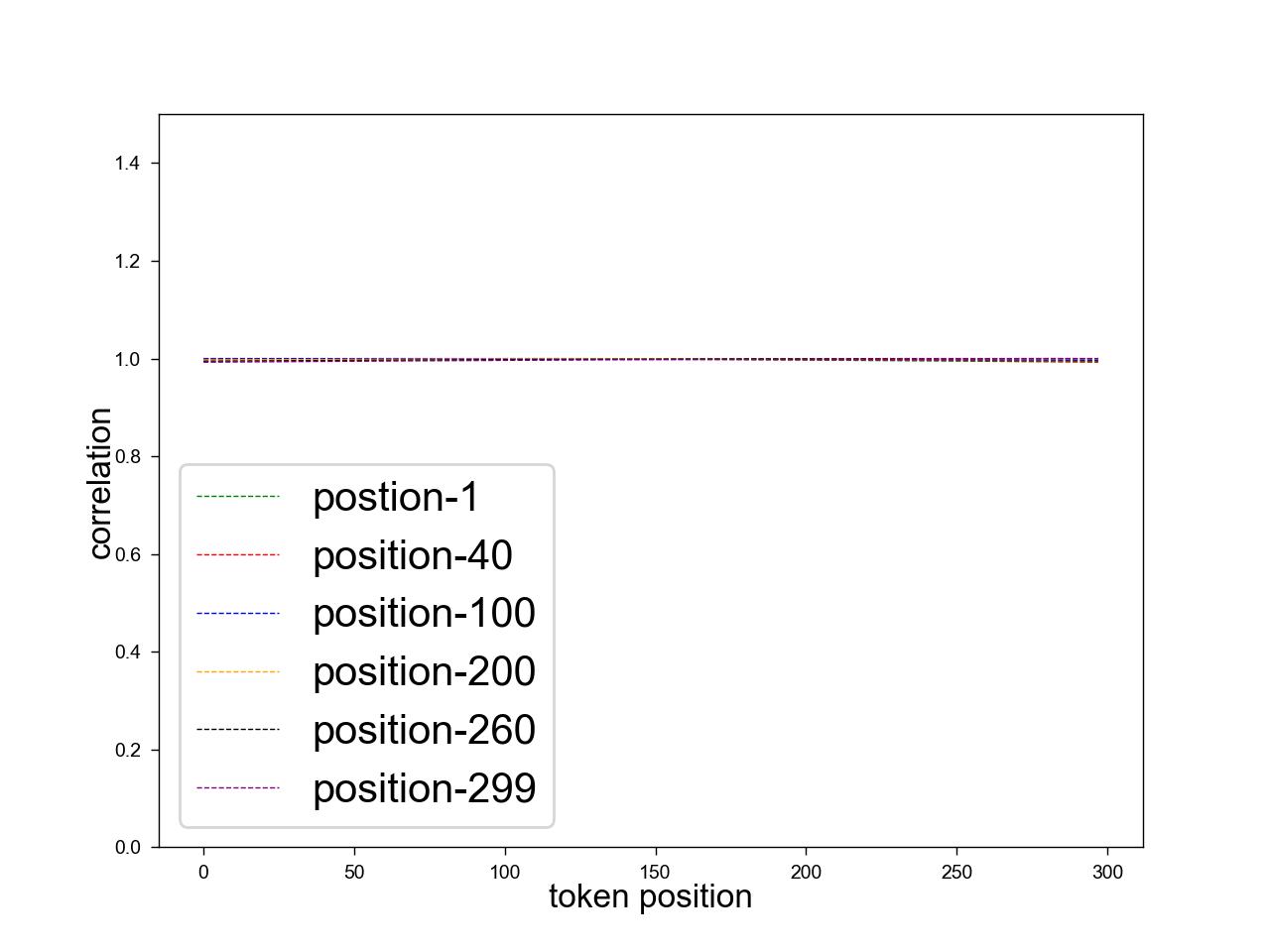}
    }
    \caption{Low variance at higher dimensional values of sinusoidal positional encoding (a), and near perfect correlation among encoding of different token positions (b)}
    \label{fig Heat Maps}
    \end{center}
\end{figure}

In the original transformer architecture, a sinusoidal position encoding scheme was proposed. The sinusoidal positional encoding method exhibits a fundamental limitation characterized by high mutual information at higher dimensions (Figure~\ref{fig Heat Maps}(a)). For instance, our analysis reveals a significant correlation trend between positional encoding values of a model with $d_{512}$ dimensions at higher dimensions ($d{>}356$) (see Figure~\ref{fig Heat Maps}(b)):
\begin{equation}
    Corr(\p_{pos=n, d > 356}, \p_{pos=m, d > 356}) 
\end{equation}
Where, $n \in \{{1, 40, 100, 200, 260, 299\}}$ and $m$ varies for all $d$\\
At higher dimensions, a near-perfect correlation (${>}0.999$ percent) is observed, even for token positions that are far apart in the sentence, such as position 299 and position 2. This observation underscores the inadequacy of sinusoidal functions in developing a dense and unique positional representation in higher dimensions when used in their original form.
In the next section, we shall demonstrate mathematically that:
\begin{enumerate}
\item This causes the model to have a biased informative prior, causing learning overheads.
\item Failing to have a proper dense representation might lead the model to lose ability to learn relative positioning effectively.
\end{enumerate}

% %%%%%%%%%% HEAT MAPS, RICHNESS OF LEGENDRE POLYNOMIALS  

%%%%%%%%%%%%%%%%%%%%%%%%%%%%%%%%%%%%%%%%%%%%%%%%

\subsection{Mathematical Analysis}

%%%%%%%%%%%%%
Equation~(\ref{eq inner product attention}) admits various interpretations. \citet{DBLP:journals/corr/abs-2006-15595} proposed the possibility of a correlation between positional encoding and words by analyzing the cross-product terms between word embedding and positional encoding. Conversely, \citet{DBLP:journals/corr/abs-2006-03654} argued that relative positional information can only be modeled using the middle two terms, i.e., the cross-terms between words and positional encoding. In this section, we present a different interpretation, focusing our attention primarily on the fourth cross-product term between two positional encoding vectors.

We begin by isolating the effect of high correlation in higher dimensions on attention, as discussed in the previous section. Without loss of generality and for better explanatory clarity, we assume that positional encodings are concatenated with token embeddings, denoted as $[\x_i | \p_i]$, rather than added to them. Functionally, these two operations are similar, requiring only slight adjustments to the projection matrices $\W_{k}$, $\W_{q}$, and $\W_{v}$. We denote these modified projection matrices as ${\W}^{}_{q} \sim \W_{q}$, ${\W}^{}{k} \sim \W{k}$, and ${\W}^{`}{v} \sim \W{v}$.

%%%%%%%%%%%%%%%%

Further, we shall rewrite $[\x_i | \p_i]$ as an augmented matrix of three blocks, i.e., token embedding, lower dimension values of positional encoding, $\p_{i}^{l-}$ (below a correlation threshold of $l$), and higher dimensional values of positional encoding, $\p_{i}^{l+}$, with high degree of correlation (Equation~(\ref{eq: embedding partition})). %Correspondingly ${\W}^{`}_{t: t\in \{q,k,v\}}$ is written in augmented form of three matrices ${\W}^{`1}_{t: t\in \{q,k,v\}}$ , ${\W}^{`2}_{t: t\in \{q,k,v\}}$ ,${\W}^{`3}_{t: t\in \{q,k,v\}}$ Equation~(\ref{eq: k projection partition}, \ref{eq: q projection partition}, \ref{eq: v projection partition}).

\begin{equation}\label{eq: embedding partition}
\mathbf{s}_i = \mathbf{[}\x_i|\p_i] = [\x_i|\p_{i}^{l-}|\p_{i}^{l+}]
\end{equation} 

%\begin{equation}\label{eq: q projection partition}
%{\W}^{`}_q = [     {\W}^{`1}_q |   {\W}^{`2}_q    | {\W}^{`3}_q             ]
%\end{equation} 
%\begin{equation}\label{eq: k projection partition}
%{\W}^{`}_k = [     {\W}^{`1}_k |   {\W}^{`2}_k   | {\W}^{`3}_k            ]
%\end{equation} 
%\begin{equation}\label{eq: v projection partition}
%{\W}^{`}_v = [     {\W}^{`1}_v |   {\W}^{`2}_v    | {\W}^{`3}_v             ]
%\end{equation} 

We can refactor Equation~(\ref{fn:adtv-posi}) in terms of augmented matrices and write the inner product self attention using query and key vectors as follows:
\begin{equation}
\q_m^{\intercal}\k_n \sim    ({\W}^{`}_q \mathbf{s}_m)^\intercal ({\W}^{`}_k \mathbf{s}_n)
\end{equation}

\begin{equation}
 = \mathbf{s}_m^\intercal  {\W}^{`\intercal}_q {\W}^{`}_k \mathbf{s}_n
\end{equation} 

Using Equation~(\ref{eq: embedding partition}) and writing the product ${\W}^{`\intercal}_q {\W}^{`}_k$ as a single matrix of three blocks corresponding to $\mathbf{s}_i$.

\begin{equation}
{\W}^{`\intercal}_q {\W}^{`}_k =[ {\W}^{1}_{kq} |   {\W}^{2}_{kq}   | {\W}^{3}_{kq}           ]  ^  \intercal
\end{equation}
from which it follows 

\begin{equation}
    \q_m^{\intercal}\k_n \sim \x_m {\W}^{1}_{kq} \x_n + \p_{m}^{l-} {\W}^{2}_{kq} \p_{n}^{l-} + \p_{m}^{l+} {\W}^{3}_{kq} \p_{n}^{l+}
\end{equation}\label{eq: cross bias term}

As discussed in the previous section, a notable degree of correlation exists between $\p_{m}^{l+}$ and $\p_{n}^{l+}$, the third term representing the generalized inner product between higher-dimensional values of positional encoding. This correlation essentially introduces a biased informative prior into the self-attention mechanism. Here, we contend that this poor representation leads to training and learning overheads. Furthermore, this effect extends to cross-terms involving word embeddings and positional embeddings, thereby hindering the learning of relative positioning as proposed by some studies (\citealp{DBLP:journals/corr/abs-2006-15595}). We conjecture the following regarding the importance of a proper dense representation of positional information:

\begin{itemize}
\item We argue that if position encoding is represented densely by some functional form, the generalized inner product between position encoding ($\p_{m}^{l+} {\W}^{3}_{kq} \p_{n}^{l+}$) has sufficient mathematical structure to it to learn relative position information. This assertion holds true for the fourth term of Equation~(\ref{eq inner product attention}).
\item While some existing methods, such as RoPE, are inherently multiplicative in nature, thereby avoid introducing an explicit bias, our analysis demonstrates that they still fail to mitigate the high degree of correlation at higher dimensions induced by sinusoidal functions used as multiplicative factors.
\item Previous works have attempted to replace positional encoding with a scalar learnable bias \citet{JMLR:v21:20-074}. In the current context, this approach resembles Equation~(\ref{eq: learnable bias}). Despite empirical and theoretical evidence suggesting the efficacy of this positional encoding scheme, it falls short in explicitly providing relative positioning information.
\item We argue that there is no absolute positional encoding scheme that satisfies all the criteria of: 1) effectively representing positional information in a dense and well-behaved manner, even in higher dimensions, and 2) encoding relative position information effectively.

\end{itemize}

\begin{equation}\label{eq: learnable bias}
    \q_m^{\intercal}\k_n \sim \x_m {\W}^{1}_{kq} \x_n + b_{mn}
\end{equation}

\section{Proposed Method: PoPE}
In this section, we propose an Orthogonal Polynomial Based Positional Encoding (PoPE) scheme. Specifically, we introduce a unique class of orthogonal polynomials referred to as Legendre polynomials. We will detail our proposed formulation of the positional encoding scheme, highlighting the desirable mathematical structure and properties of Legendre polynomials that address certain limitations observed in existing positional encoding schemes.

\subsection{Legendre Polynomials}

\begin{figure}[h]
    \begin{center}
    \includegraphics[height=5cm]{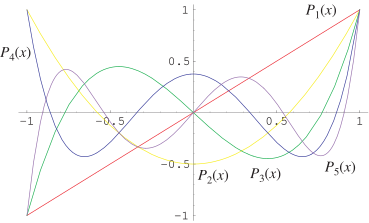}
    \caption{First four order Legendre Polynomials, credit: $Mathworld$}
    \label{fig: First four order Legendre Polynomials}
    \end{center}
\end{figure}

Legendre polynomials $\P_n(x)$ are special category of orthogonal polynomials (Figure~\ref{fig: First four order Legendre Polynomials}). They appear in numerous engineering and physics problems, especially related to spherical harmonics (\citealp{Dassios_2012}). One of the compact forms of Legendre polynomials is given by Rodrigues' formula (Equation~(\ref{eq rodrigues})).
%%%%%%%%%%% LEGENDRE POLYNOMIAL GRAPHS, EXAMPLES 
%%%%%%%%%%%%%%%%%%%%%%%%%%%%%%%%%%%%%%%%%%%%%%%%%%%

\begin{align}\label{eq rodrigues}
 \textbf{P}_{n}(x) = \frac{1}{2^n n!}\frac{d^n }{dx^n}(x^2-1)^n
\end{align}
Where, $\mathbf{P}_{n}(x)$ is Legendre polynomial of order n.
Legendre polynomials are orthogonal functions under the restricted domain  $x \in [-1, 1]$ (Equation~(\ref{eq orthogonality of Pmn})), thus, just as sinusoidal functions, can be chosen as basis functions in Fourier system.

\begin{equation}\label{eq orthogonality of Pmn}
    \int_{-1}^{1} \P_{n}(x)\P_{m}(x) \,dx = \delta_{mn}
\end{equation}

\subsection{Proposed Formulation}
For a token of Position $= pos$, we propose deterministic and equidistant sampling from a Legendre polynomial $\P_n(x)$ ($x \in [-1, 1]$) of order $n = pos$, hence for a given $d_{model}$, Equation~(\ref{eq PoPE Scheme}) defines the value of position encoding for a given position $pos$ and index $i$.

\begin{equation}\label{eq PoPE Scheme}
    PE_{(pos,i)} = \mathbf{P}_{pos}(x_{i})
\end{equation}

Where, $x_{i}$ are equal interval samples in $[-1, 1]$ with interval length $ = 2i/d_{model}$.

\subsection{Properties of PoPE}

\begin{figure}[H]
    \begin{center}
    \subfigure[PoPE Scheme]{
    \includegraphics[height=5cm,width=0.5\textwidth,scale=1]{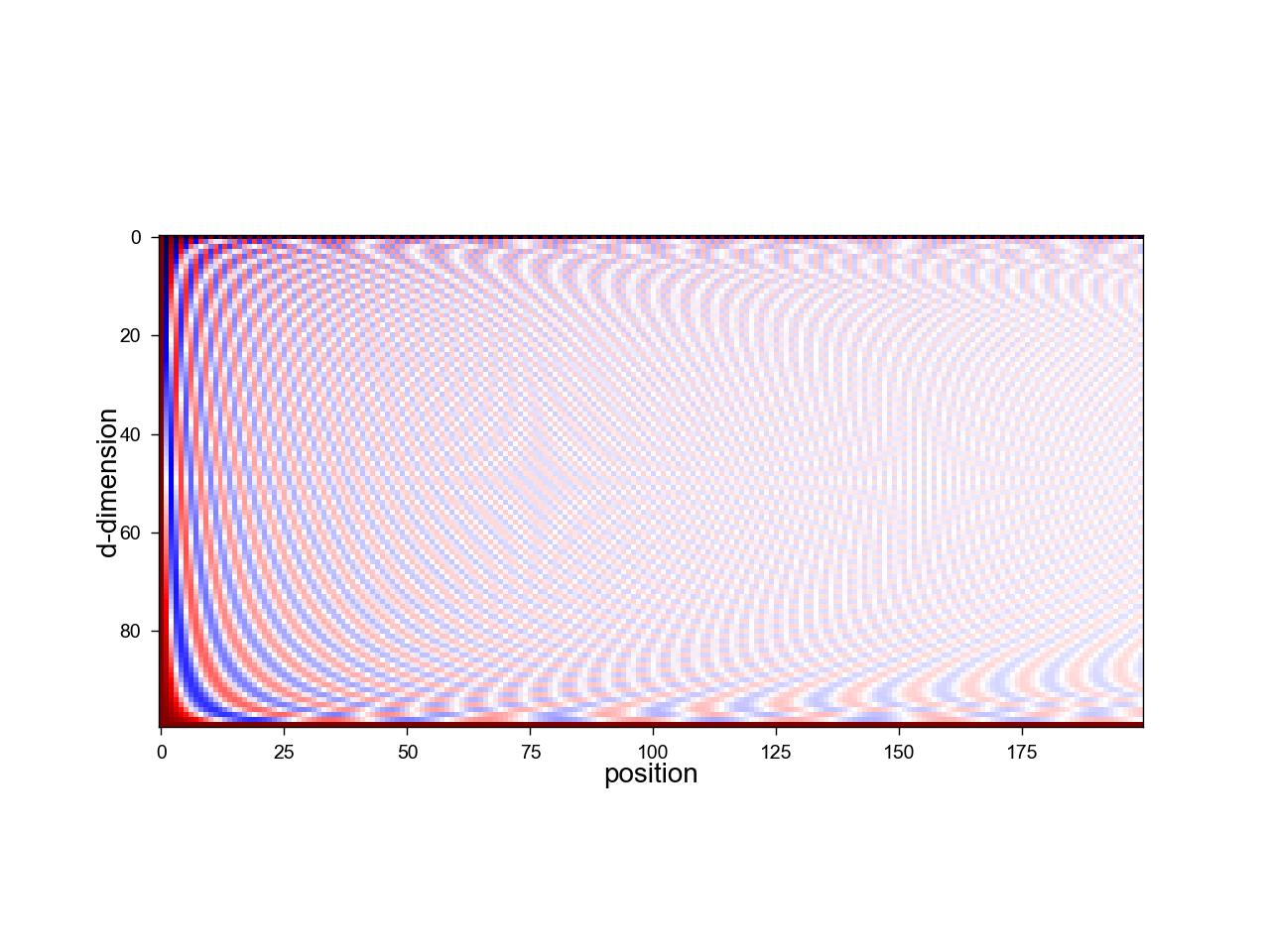}
    }
    \subfigure[Much better correlation structure at higher dimensions]{
    \includegraphics[height=5cm, scale=1,width=0.5\textwidth]{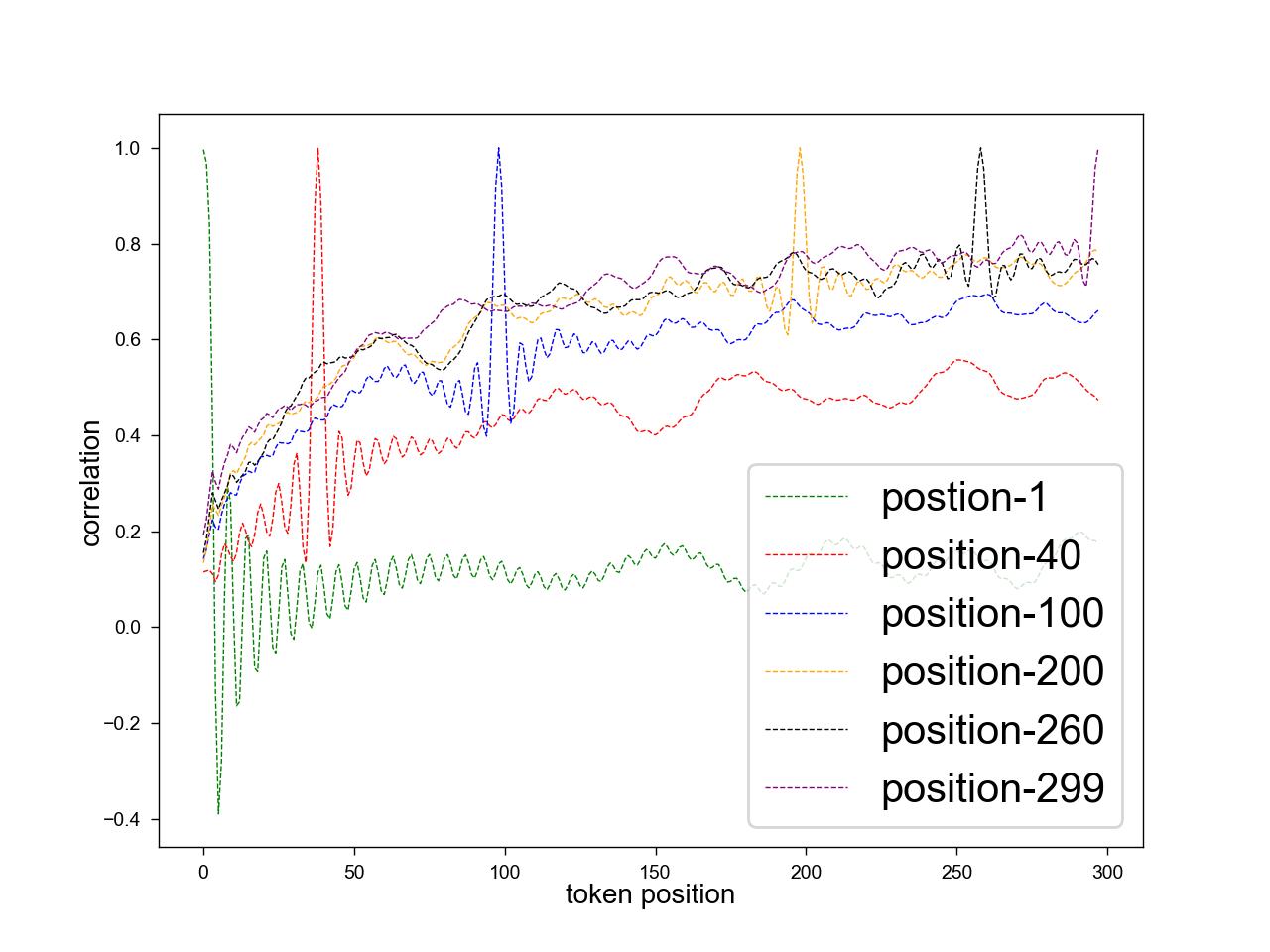}
    }
    \caption{PoPE has much better representation (a), correlation among token positions are much better managed  (b)}
    \label{fig poly heat maps}
    \end{center}
\end{figure}

\textbf{Well behaved at higher dimensions:}
Similar to sinusoidal functions, Legendre polynomials exhibit orthogonality; however, they possess non-periodic characteristics, with each polynomial of order $n$ having a distinct functional form. Drawing an analogy from signal processing, sinusoidal position encoding resembles frequency or phase modulation, where position is encoded by altering the phase or frequency (both represented as angular terms in a sinusoidal function). In contrast, Legendre polynomials can be likened to modulation in both phase and amplitude. This dual modulation results in increased entropy, allowing for the encoding of a greater amount of information even in higher dimensions. Consequently, Legendre polynomials offer a more comprehensive and efficient representation  (Figure~\ref{fig poly heat maps}(a)).
Improved dense representation offers two significant advantages. Firstly, the cross-product term between positional encoding will not exhibit a biased informative prior. As illustrated in Figure~ \ref{fig poly heat maps}(b), the correlation is more accurately represented as a function of distance. Additionally, as previously argued, enhanced representation allows the generalized inner product term between position encoding to effectively learn relative positioning between tokens. This effect will be further pronounced due to other properties of PoPE, which we will discuss in the following section.

% %%%%%%%%%% HEAT MAPS, RICHNESS OF LEGENDRE POLYNOMIALS  

%%%%%%%%%%%%%%%%%%%%%%%%%%%%%%%%%%%%%%%%%%%%%%%%

\textbf{Learning relative positions:}
It is often considered beneficial for positional encoding to have a learnable linear structure between positions, which means for any given offset $k$ positional encoding $PE_{pos+k}$ can be represented as a linear function of $PE_{pos}$. The original transformer paper (\citealp{DBLP:journals/corr/VaswaniSPUJGKP17}) claims this for sinusoidal positional encoding. In this section we shall show how PoPE has a well defined linear recurrent structure which helps us to define relative positions.
By virtue of Favard's theorem (\citealp{zbMATH02532007}), all orthogonal polynomials satisfy a three-recurrence relations, Equation~(\ref{eq: legendre 3-recurrence}) defines a three-recurrence relation for a Legendre polynomial.

\begin{equation} 
    (2n + 1)xP_{n}(x) = nP_{n-1}(x) + (n + 1)P_{n+1}(x)\label{eq: legendre 3-recurrence}
\end{equation}

This recurrent relation for Legendre polynomials can be formulated as a tri-diagonal Jacobi matrix (\citealp{VANASSCHE2022133214}), refer Equation~(\ref{jacobi}). The Jacobi matrix truncated to any order $n$ has eigen values equal to the roots of Legendre polynomial of order $n$, this implies relation recurrence relations between Legendre polynomials contain complete information about any $P_{n}(x)$
It is interesting to note that Legendre polynomial of any degree has a linear relation with one lower and higher order polynomial, which can be extended indefinitely over any order $n$ $[0, \infty )$. Also to be noted that for a given model architecture, $X$ is a constant, since it is fixed $d_{model}$ sample values from interval $[-1, 1]$, where $d_{model}$ corresponds to the dimension of the token embedding.
This is a highly desirable property of Legendre polynomials and orthogonal polynomials in general, it provides a firm mathematical structure to each position encoding of a token in a sentence with respect to each position in both forward and backward direction.

\begin{align}
\mathbf{X}
\begin{bmatrix}
    P_{0}(x) \\ P_{1}(x) \\ P_{2}(x) \\ P_{3}(x) \\ \vdots \\ P_{n-1}(x) \\ P_{n}(x)
\end{bmatrix} = 
\begin{bmatrix}
    b_{0} & a_{0} & 0 & 0 & 0 & \dots & 0\\
    c_{1} & b_{1} & a_{1} & 0 & 0 & \dots & 0 \\
    0 & c_{2} & b_{2}& a_{2} & 0 & \dots & 0 \\
    0 & 0 & c_{3} & b_{3} & a_{3} & \dots & 0 \\
    \vdots & \vdots & \vdots & \vdots & \vdots & \ddots & 0 \\
     \vdots & \vdots & \vdots & c_{n-1} & b_{n-1} & a_{n-1} & 0 \\
    \dots & \dots & \dots & \dots & c_{n} & b_{n} & a_{n}
\end{bmatrix}
\begin{bmatrix}
    P_{0}(x) \\ P_{1}(x) \\ P_{2}(x) \\ P_{3}(x) \\ \vdots \\ P_{n-1}(x) \\ P_{n}(x)
\end{bmatrix} 
\label{jacobi}
\end{align}

Hence, any Legendre polynomial can be expressed as a linear combination of other Legendre polynomials of higher and lower orders through nested recursion relations. This indicates that the position encoding for any given token position inherently contains relative information from all other positions embedded within it, owing to the recurrence relation.\\
$ \P_n(x) = \sum_{l} a_{l}\P_{n+l}(x) $\\
Where, $l \in \{[n, \infty) - \{0\}\}$\\
We argue that this property facilitates the injection of relative position information into the model, thereby enhancing the model's capability to learn relative position information through generalized inner product terms between positional encodings ($\p_{m}^{l+/-} {\W}^{3}_{kq} \p_{n}^{l+/-}$ and $\p_m^{\intercal}\W_q^{\intercal}\W_k\p_n$).
The inner product term between position encoding of $m^{th}$ and $n^{th}$ token diffuses across position encoding values, thereby imparting relative position information to the model.\\
 $\p_m^{\intercal}\W_q^{\intercal}\W_k\p_n \sim \sum_{l}  \sum_{j} a_{i}a_{j}\p_i^{\intercal}\W_q^{\intercal}\W_k\p_j $ \\
 Where $a_{i}, a_{j}$ are specific to any given two position $m$ and $n$.
\section{Experiments and Results}
We assessed the performance of the baseline transformer augmented with PoPE on the $Multi30K$ English-to-German translation task. In this section, we provide details on the experimental setup, including hardware specifications, model parameters, and the comparative analysis of PoPE with previous benchmarks.
\subsection{Data and task}
We evaluated the performance of the base transformer model with both sinusoidal and Legendre polynomial-based positional encoding (PoPE) schemes on a bilingual (English to German) language translation task using the $Multi30K$ dataset (\citealp{DBLP:journals/corr/ElliottFSS16}). $Multi30K$ dataset extends the $Flickr30K$ dataset with translated and independent German sentences (\citealp{young-etal-2014-image}). Dataset contains 29,000 training and 1,014 validation sentence pairs. There are three test datasets available $test\_2016\_flickr$ and $test\_2017\_flickr$ with 1,000 sentence pairs, and $test\_2017\_mscoco$ with 461 sentences. In this paper we have used all testing samples.
Due to limited compute resources at disposal, $Multi30K$ provided the ideal experiment data to experimentally support theoretical and empirical merits and aspects of PoPE discussed in this paper. 

\subsection{Model and Hardware}

We carried out the experimentation using the base transformer model, as defined in the original transformer with $d_{model} = 512$, eight attention heads $h=8$, and $d_q = d_k = d_v = d_{model}/h$, dimensions of inner layers $d_{ff}=2048$, six encoder and decoder blocks $N =6$.
Due to compute restrictions, we did not evaluate the transformer (big) model and larger WMT 2014 dataset, this is a work in progress. We trained and evaluated the models on 1 NVIDIA RTX™ A4500 GPU. To make a one to one comparison, we trained the model for 10,000 iterations on $Multi30k$ dataset, inline with \citet{zhang2019neural}.
\subsection{Results}
On $Multi30k$ English to German translation task, the base transformer model with PoPE achieves a BLEU score of 40.7, which is 4-5 BLEU more than base and big transformer models with sinusoidal positional encoding, achieving SOTA performance among baseline transformer models and establishing new a new benchmark, it is to be noted that, while $Multi30k$ is a multimodal dataset, we are comparing NMT text to text performance here. There are many multimodal studies which outperform text to text models on $Multi30K$ English to German translation task (IKD-MMT (\citealp{peng-etal-2022-distill}) BLEU 41.28, ERNIE-UniX2 BLEU 49.3 (\citealp{shan2022ernieunix2})), but base transformer with PoPE was able to outperform some of the multimodal approaches (DCCN (\citealp{10.1145/3394171.3413715}) BLEU 39.7, Caglayan (\citealp{caglayan}) BLEU 39.4, Multimodal Transformer (\citealp{yao-wan-2020-multimodal}) BLEU 38.7). \\

\begin{figure}[H]
    \begin{center}
    \includegraphics[height=8cm]{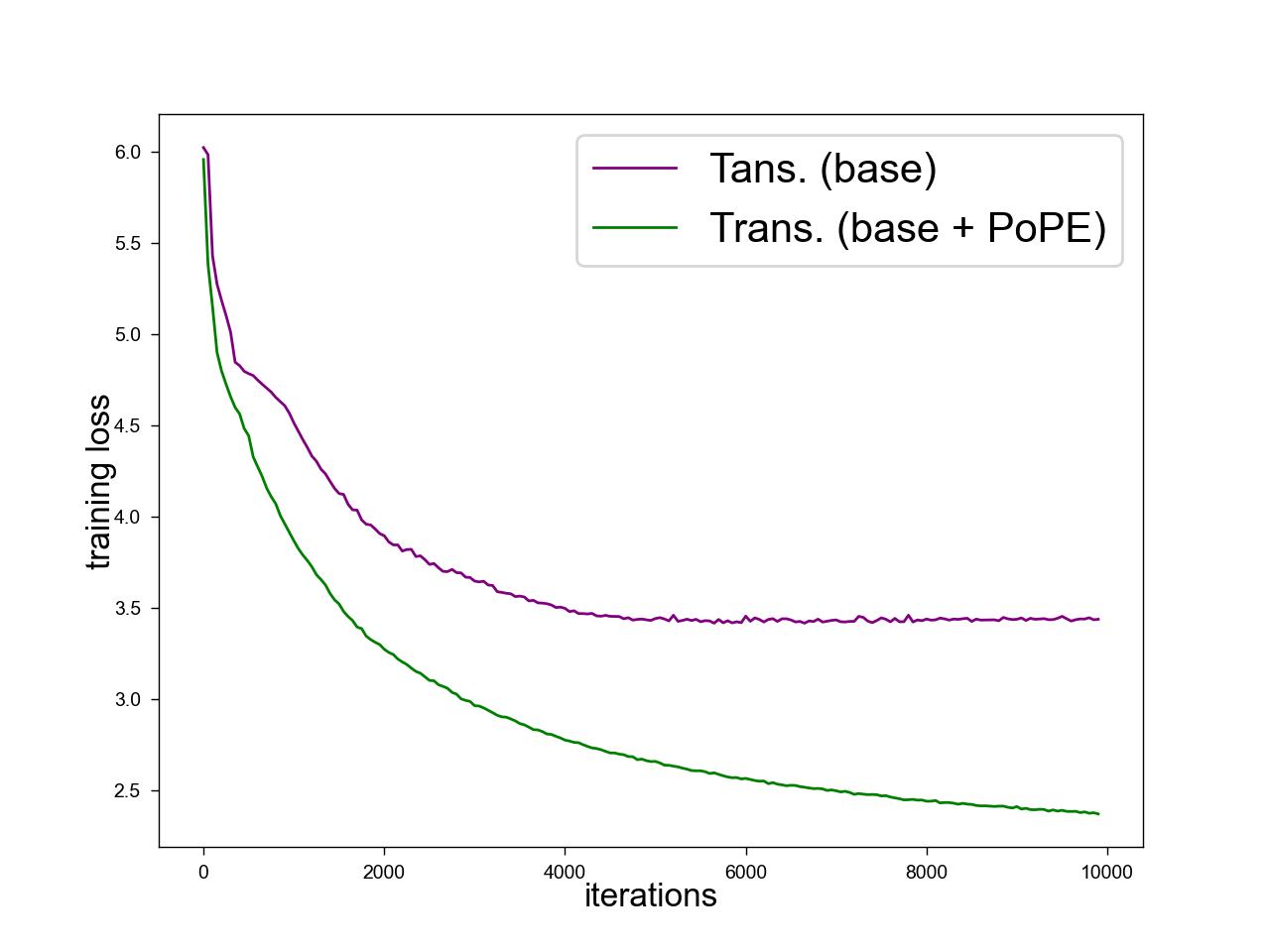}
    \caption{Training loss convergecne with and wihout PoPE}
    \label{fig: loss convergence}
    \end{center}
\end{figure}

In our training routines, the proposed model (transformer + PoPE) converged at two to three times faster than original transformer model with sinusoidal positional encoding (Figure~\ref{fig: loss convergence}). 

\begin{table}[h]
\caption{Comparison with Transformer Based Text to Text NMT on EN-DE Task}
\centering
\begin{tabular}{@{}cccccc@{}}
\hline
\textbf{Study} & \textbf{Model} &\textbf{Task} & \textbf{Dataset}& \textbf{Training}& \textbf{BLEU} \\
&&&& \textbf{steps} & \textbf{score}\\
\hline
\citet{DBLP:journals/corr/VaswaniSPUJGKP17}&Transformer (base) & EN to DE & $WMT$ 2014 & 100,000 & 27.3\\
&Transformer (big) & EN to DE & $WMT$ 2014 & 300,000 & 28.4\\
\hline
\citet{zhang2019neural}&Transformer (base) & EN to DE & $Multi30k$ & 10,000 & 35.59\\
&Transformer (big) & EN to DE & $Multi30k$ & 10,000 & 36.86\\
\hline
This study (PoPE) & Transformer (base) & EN to DE & $Multi30k$ & 10,000 & \textbf{40.7}\\
\hline
\end{tabular}\label{tab Benchmarks}
\end{table}

%%%%%%%%%%% LEGENDRE POLYNOMIAL GRAPHS, EXAMPLES 
%%%%%%%%%%%%%%%%%%%%%%%%%%%%%%%%%%%%%%%%%%%%%%%%%%%

\section{Discussion}
Through empirical, theoretical, and experimental analyses presented in this paper, we have demonstrated how better representation can address many issues related to positional encoding. We showcased the advantages of Orthogonal Polynomial-based Positional Encoding (PoPE), highlighting its desirable mathematical properties and superior performance in comparison to other models, with significantly improved convergence rates. \\
In summary, our this paper yields several key insights:
\begin{enumerate}
    \item PoPE offers superior representation of positional encoding even at higher dimensions compared to sinusoidal positional encoding, which often exhibits highly correlated values at higher dimensions. This can be attributed to limited encoding capability of sinusoidal functions with frequency/phase modulation alone at higher dimensions, various propertied of Legendre polynomials such as orthogonality, non periodicity, and different functional form of the Legendre polynomial functions of different orders within $[-1, 1]$, which proves them to be much better basis than sinusoidal functions for encoding position information effectively. It is easy to prove that even rotatory encoding scheme does not solve this problem, though multiplication avoids a bias term in attention and prove to be marginally better than APE method using sinusoidal functions.
    \item From theory it is indicative that that the cross term between higher dimension values of position encoding  ($\p_{m}^{l+} {\W}^{3}_{kq} \p_{n}^{l+}$)  behave like a biased informative prior, and is an additional learning overhead. We argue this induces bias can potentially cause slow convergence of the model. PoPE avoids this problem altogether by better representation and thus have much superior converging properties, proven by experiment results. This observation sheds light on why RoPE-based positional encoding, which explicitly avoids a bias term, exhibits slightly faster convergence compared to baseline transformer models, a phenomenon noted but not fully addressed in the original RoPE paper (\citealp{DBLP:journals/corr/abs-2104-09864}). From this, it is evident that there is  some merit in avoiding a weak positional encoding altogether and explicitly learn a scaler valued bias (\citealp{JMLR:v21:20-074}).

    \item We argue that providing relative position information to the model offers little merit. We argue, with dense representation across dimensions, the generalized inner product term($ \p_{n}^{l+/-}{\W}^{3}_{kq}  \p_{m}^{l+/-}$) is sufficient to infer relative positioning. Methods such as RoPE and are categorized as relative positional encoding scheme, but performs much like absolute positional encoding schemes (\citealp{NEURIPS2023_4e85362c}) with marginal benefits (\citealp{shaw2018selfattention}).

    \item Theoretically, PoPE-based encoding schemes can learn both absolute and relative position information effectively due to their better representation and the existence of three-recursion relations among orthogonal polynomials. The linear relation between Legendre polynomials of different orders (and hence positions) is more comprehensive than that afforded by sinusoidal functions.

\end{enumerate}

\section{Limitations and Future Scope}

\begin{itemize} 
\item
We have tested the proposed PoPE scheme on a smaller $Multi30K$ dataset. Further work is needed to gather more experimental evidence on much larger benchmark datasets and task to fully assess the capability of PoPE.
\item
Merits of PoPE are not restricted to the structure of the original transformer paper, it could also be used in other modeling paradigms as well, such as providing per head positional encoding, which removes rank restriction on attention matrices (\citealp{chen-etal-2021-simple}). Further work is required to explore full potential of PoPE.
\item 
We have demonstrated that the recurrence properties of Legendre polynomials possess sufficient structure, akin to sinusoidal functions, to facilitate the linear learning of relative positions. However, the relationship becomes more intricate with orthogonal polynomials. Further research is required to comprehensively develop a mathematical understanding of how these complex relations can aid large language models such as transformers.

\end{itemize}

\section{Declaration}
The views expressed in this paper are solely those of the author.

\vskip 0.2in
\bibliography{main}

\end{document}